\documentclass[11pt]{article}
\usepackage[ backend=biber,style=numeric,sorting=none ]{biblatex}
\usepackage{graphicx} 
\usepackage{amsmath, amsthm, amssymb}
\usepackage{makecell}
\usepackage{hyperref}
\usepackage{setspace}
\usepackage{float}
\usepackage{multicol}
\usepackage[margin=1in]{geometry}
\usepackage{pdflscape}
\usepackage{nicematrix}
\usepackage[font=small, labelfont=bf]{caption}
\usepackage{afterpage}
\usepackage{authblk}
\graphicspath{./images/}

\title{Mapping waterways worldwide with deep learning}
\author[1,2,3,*]{Matthew Pierson}
\author[1,2,3,*]{Zia Mehrabi}
\affil[1]{Better Planet Laboratory, University of Colorado Boulder, Boulder, 80309, Colorado, USA}
\affil[2]{Department of Environmental Studies, University of Colorado Boulder, Boulder, 80309, Colorado, USA}
\affil[3]{Mortenson Center for Global Engineering and Resilience, University of Colorado Boulder, Boulder,8 0309, Colorado, USA}
\affil[*]{Corresponding authors. E-mails: \href{mailto:Matthew.Pierson@Colorado.edu}{Matthew.Pierson@Colorado.edu}, \href{mailto:zia.mehrabi@colorado.edu}{Zia.Mehrabi@Colorado.edu} } 

\begin{document}

\maketitle


\newcommand{\precision}{0.7200}
\newcommand{\recall}{0.6034}
\newcommand{\fone}{0.6566}

\newcommand{\precisionignore}{0.8235}
\newcommand{\recallignore}{0.6446}
\newcommand{\foneignore}{0.7232}

\newcommand{\precisionmask}{0.6888}
\newcommand{\recallmask}{0.7236}
\newcommand{\fonemask}{0.7058}

\newcommand{\tanimoto}{0.5}
\newcommand{\learningrate}{0.0625}

\newcommand{\E}{\text{Elevation}}

\begin{abstract}
    Waterways shape earth system processes and human societies, and a better understanding of their distribution can assist in a range of applications from earth system modeling to human development and disaster response.  Most efforts to date to map the world's waterways have required extensive modeling and contextual expert input, and are costly to repeat. Many gaps remain, particularly in geographies with lower economic development. Here we present a computer vision model that can draw waterways based on 10m Sentinel-2 satellite imagery and the 30m GLO-30 Copernicus digital elevation model, trained using high fidelity waterways data from the United States.  We couple this model with a vectorization process to map waterways worldwide. For widespread utility and downstream modelling efforts, we scaffold this new data on the backbone of existing mapped basins and waterways from another dataset, TDX-Hydro. In total, we add ${\sim}$124 million kilometers of waterways to the ${\sim}$54  million kilometers already in the TDX-Hydro dataset, more than tripling the extent of waterways mapped globally.
\end{abstract}

\section{Introduction}

Many waterways around the world don't appear in easily accessible geospatial datasets, such as Open Street Map (OSM) \cite{OpenStreetMap}. Even recently developed global waterways datasets, such as TDX-Hydro \cite{carlson_2024_tdxhydro}, created by the United States National Geospatial Intelligence Agency using a high resolution Digital Elevation Model (12m TanDEM-X), are missing many small tributaries. This is problematic for a number of applications, including, but not limited to rural infrastructure development projects. For example, in our recent work investigating the impact of rural trail bridges on access to schools, health care facilities, and markets across different countries in Africa we found many cases where communities would state a need for a bridge in places where no waterway were mapped \cite{Pierson2024_1}. And while we have found that using TDX-Hydro fills in many of the missing waterways in Open Street Map, even this state of the art data misses a substantial quantity of community bridge requests \cite{Pierson2024_1}. There is a clear need for a more comprehensive dataset of waterways across the world.

In our previous work, we used a machine learning model, WaterNet, to map waterways based on the National Hydrography Dataset from the USA, trained using 10m satellite Sentinel-2 Level-2A NRGB bands and the 30m GLO-30 Copernicus DEM . We deployed this model in 8 European and 8 African countries, and found we were not only able to reproduce results from independently developed TDX-Hydro, but were able to capture more community infrastructure needs of rural populations in Africa \cite{Pierson2024_1}. This was our first indication that a combination of computer vision, with high resolution satellite imagery, could hold significant promise for developing waterways data in a cost-efficient and scalable way, across large swathes of the Earths surface.

In our previous work, we restricted the geographic scope of deployment of WaterNet.  We did not evaluate its value globally for mapping waterways relative to other existing data.  In this paper, we report a global scale inference of this model for every continent and most large islands across the world.  We also report the training process, architecture, and inner workings of the model, as well as updates to our previous work. One of the key updates includes utilizing the backbone of TDX-Hydro in our vectorization process, which we do for maximum consistency with existing data, to assess the value addition against this known benchmark, and to enable easier interoperability with existing applications. We assess the additional waterways that are added by WaterNet to this backbone, and publicly release both the accompanying raster and vector data, alongside the model, and code, to the scientific community for further research and inclusion in their downstream tasks \cite{YY2XMG_2024}.

\section{Results}

\subsection{Model overview}

To interpret the results and model output it is important to understand some details of the model itself, which we briefly overview here (see Methods \ref{methods} for more detail). WaterNet is generally based on ideas from U-Net \cite{ronneberger2015unet} and ResNet \cite{he2015}. Details of the model are included in the Model Tables \ref{modtab}, including information on model layers \ref{tab:model}, convolutions \ref{tab:convolution}, decoders \ref{tab:decoder}, encoders \ref{tab:encoder},  gated linear units   \ref{tab:multiplication}, and residual layers \ref{tab:residual} - \ref{tab:residualB}. One of the unique aspects of this model is that we don't complete the U-Net. That is to say, we use 5 encoders (decreasing the width and height of each image by a factor of two at each iteration), and we only use  4 decoders, optimizing storage while maintaining precision of raster outputs that are 20m globally (higher resolution from our earlier 40m version).  These rasters are then vectorized by first connecting our waterways to the TDX-Hydro waterways using least cost pathing to connect disconnected segments, on top of which we employ a thinning and vectorization algorithm.

Notably, the model was trained across a diversity of hydrographic conditions using labels from the National Hydrography Dataset  (e.g. with a waterways identifier for each water type such as rivers, streams, lakes, ditches, intermittent, ephemeral). We do this in two steps, starting  with a larger training set of smaller context ${\sim}$1.5M grids (244 x 244 pixels),  and followed with a ${\sim}$10x decrease in training samples but ${\sim}$ 10x increase in context, ${\sim}$90K grids (832 x 832 pixels). We have found this two step approach to be a useful for making location predictions across a diversity of contexts and water way types, while at the same time minimizing evaluation time and maximizing speed and alignment of waterways network structures in the final product. We use a summed Binary Cross Entropy and Tanimoto loss weighted by waterway type (Supplementary Table \ref{tab:edt2}) . We effectively mask swamps, canals, intermittent lakes, ditches, and playas in training, with rivers and streams, intermittent, ephemeral and perennial, alongside perennial and permanent lakes being our primary target  - although we evaluate the model performance on all waterway types, see below. Our input features includes 10 channels: the first four being transformed Sentinel-2 NRGB channels ($NRGB_t$), and the remaining 7 being $NDVI$, $NDWI$, Shifted Elevation ($E_S$), Elevation x-delta ($\Delta_x E$), Elevation y-delta ($\Delta_y E$), elevation gradient ($\nabla E$).
 
\subsection{Performance}

\begin{landscape}
\begin{figure}
    \centering
    \includegraphics[width=9in]{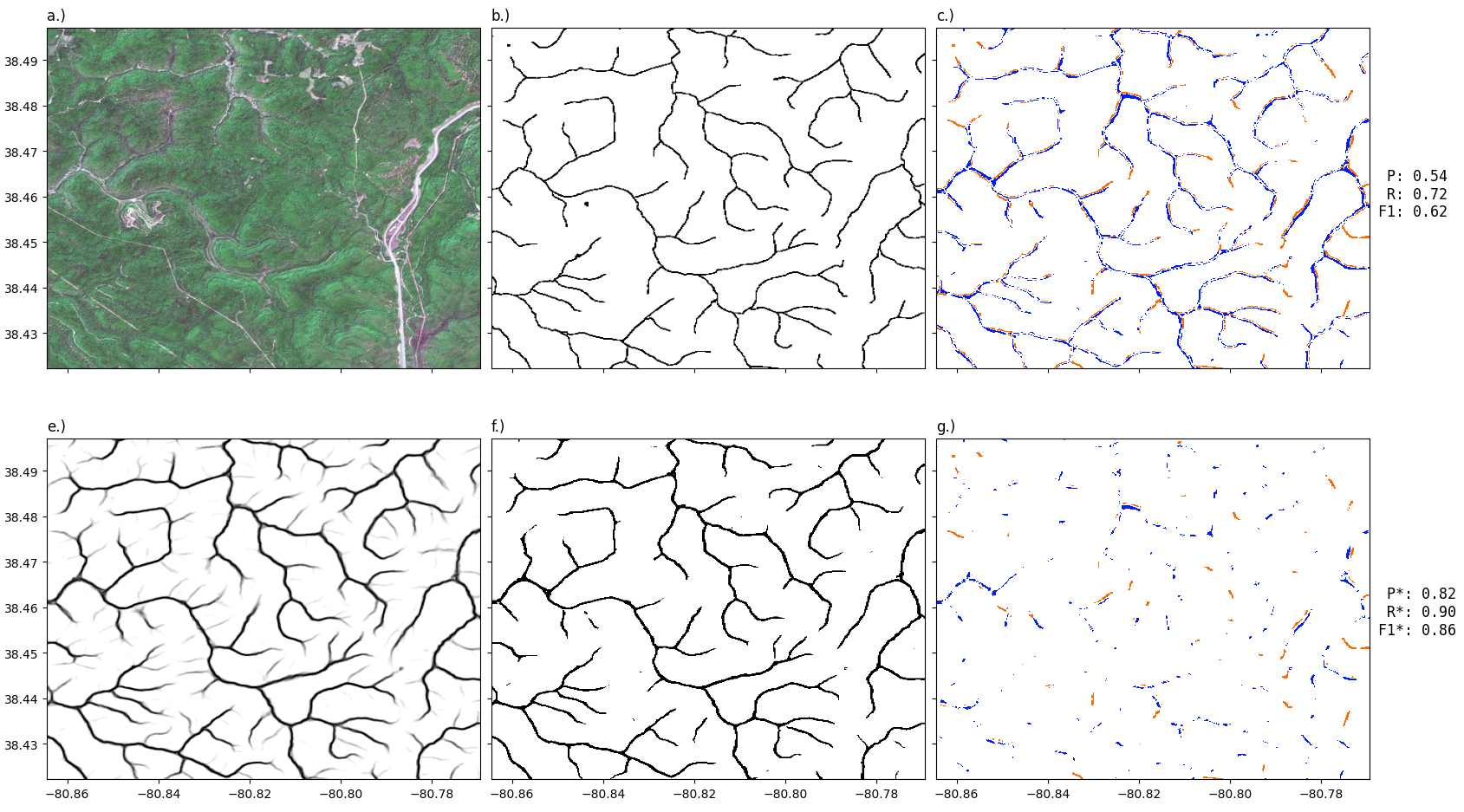}
    \caption{\textbf{WaterNet in action.} a) Input Sentinel-2 data, b) Burned National Hydrography Data (NHD), e) Model output, f) Model output rounded at a probability of 0.5, c) Difference between the rounded output and the NHD data, g) Difference with line thickness tolerance.  This highlights how pixel precision (P), recall (R), and F1 are greatly affected by the model's predictions being too thick. We find that the precision (0.82), recall (0.9), and F1 (0.86) found when removing the effect of waterway thickness in raster outputs offers a better representation of the model's ability to pick up network structures than the same metrics on raw raster outputs (P=0.54, R=0.72, F1=0.62). Blue = false positive, Orange = false negative}
    \label{fig:thick}
\end{figure}
\end{landscape}

Waterways, like roads \cite{Microsoftroads2021}, require special attention to accuracy assessment. We computed accuracy statistics using bespoke test statistics, which are required to fully understand the performance of the model (see  Table~\ref{tab:scores}).  
The first of these computes pixel level precision ($P^*=\precisionignore$), recall ($R^*=\recallignore$),
and F1 score ($F1^*=\foneignore$), that do not penalize for minor errors in line thickness. These are our preferable test statistics, because the commonly used, or standard baseline  ($P=\precision$), recall $(R=\recall)$, and F1 ($F1=\fone$) are contaminated by a large number of mismatches from line width differences, despite the model representing the waterways spatial pattern with high fidelity. An example can be seen in Figure~\ref{fig:thick}, and by the difference between these statistics ($P^*-P\approx 0.1235,\, R^*-R\approx 0.0412$, \, $F1^*-F1\approx 0.0666$).

We additionally explore model performance by both creating mask specific test statistics and through experiments on accuracy by the waterway types effectively masked during training.
That is we computed the ($P^{**}=\precisionmask$), recall ($R^{**}=\recallmask$),
and F1 score ($F1^{**}=\fonemask$) which ignores pixels that had waterway types that were masked out during training. We find that while the precision decreases, which is due to decreasing the number of true positives without affecting the number of false positives, the recall and F1 increase greatly. 
($P^{**}-P\approx -0.032,\, R^{**}-R\approx 0.1202$, \, $F1^{**}-F1\approx 0.0492$). An  example of this can be seen in Figure~\ref{fig:mask}, which shows why including masked labels in the test set leads to low recall. Experiments on test data subsets show  that a key source of inaccuracy arises from prediction of intermittent lakes, we see boosts in F1 score when these are removed from the test set (see Table~\ref{tab:scores}). While these test statistics are useful for diagnostics, we  recommend the use $P^*$ , $R^*$ , $F1^*$ more generally for people using this model for downstream tasks where accuracy of all  (even those masked during training) waterway structures are important.

\subsection{Global deployment}

Driven by the performance of WaterNet across a wide range of hydrographic conditions across the United States, alongside prior performance in mapping waterways in Africa \cite{Pierson2024_1}, we set out to deploy it globally. This required processing all 10 input channels for 10m Sentinel-2 Level-2A NRGB bands to mosaic a cloud free images for circa 2023, and conducting inference. A raster output of the global extent of our predictions made at 20m are shown in  Figure~\ref{fig:global}. Regional examples taken from this global raster layer are shown in  Figure~\ref{fig:regional}, which illustrates major waterways and associated tributaries for the Mississippi, Amazon, and the Congo river basins. 

While this data set is circa 2023, the global inference time and post processing time for the raster output is 9 days, and vectorization output 10 days, on a modest machine with Intel I9 13900K 24 core CPU Nvidia 3090Ti and 128GB Memory (and due to parallelization, these are speeds that would scale with additional GPU and CPUs, respectively). Recreating a global map of waterways using this method is thus highly amendable to repeat predictions which leverage new satellite imagery, which make it particularly useful for operational contexts.  We also distribute  global vectorized version of this data on the backbone of the TDX-Hydro data vectorized each Level 2 basin in the HydroBASINS dataset \cite{Lehner} for use in downstream applications (see Methods \ref{methods}).

\subsection{Total waterways added}

In total we add 132,986,677 kilometers of waterway to the 58,593,547 kilometers already in the TDX-Hydro dataset. When removing artifacts from vectorization (which can add multiple lines for each lake, for example) we find  we add 124,678,321 kilometers of waterway to the 54,950,267 already in TDX-Hydro. We further calculated these added lengths filtering by stream order (Table ~\ref{tab:lenadded}). The significant gains arise from order 1, 2 and 3 streams, representing more than 75M, 38M and 10M new waterways mapped, respectively. Breakdowns by basin are also given in Supplementary Table~\ref{tab:lenbybasin}. These gains are substantial and significantly changes our understanding of the distribution of waterways across the world. 

\subsection{Type of waterway added}

A critical question that remains is why does WaterNet capture so many more waterways than any other existing datasets? Insights from our prior work in rural Africa were useful here, where we found waterways can provide an obstruction to human movement, that is to the extent communities will request a trail bridge to cross, and that this obstruction can be important, even if it exists for a short time period in the year \cite{Pierson2024_1} .  This provides a good indication WaterNet is picking up intermittent and ephemeral waterways, which have to date remained poorly mapped at high resolution. 

While there have been recent efforts to map intermittent and ephemeral waterways \cite{Messager2021-yp},  inter-comparisons with our product are limited because prior efforts fail to capture lower order streams. We were however able to assess which classes of waterway WaterNet added to existing TDX-Hydro data using existing classifications of waterways types from the NHD data in the United States. We found that while the type of new streams detected by WaterNet depended heavily on location and stream order (Supplementary Table \ref{tab:streamtype}),  a general trend was that as stream order decreased, WaterNet added  waterways that were more likely themselves ephemeral (based on nearest neighbour statistics).  We do find geographic differences, for example, with WaterNet adding more ephemeral and intermittent streams to the various desert regions in the USA and more intermittent and perennial streams to the eastern US. While it is hard to extrapolate these findings globally,  they do indicate the value of WaterNet for capturing unmapped ephemeral and intermittent streams.

\section{Discussion}

We have described a machine learning model to map waterways globally, trained using high resolution satellite imagery and a moderate resolution digital elevation model, as far as we know, the first model and dataset of it's kind at this scale. It is our understanding that this new methodology and data presents an important advance to complement existing waterways mapping efforts globally \cite{  Lehner,Allen2018,carlson_2024_tdxhydro,Pekel2016-bt}, particularly in representation of lower order and non-perennial stream structures. In total our new mapping efforts more than triple the extent of waterways mapped globally. 

Because we have, in our vectorization process, extended existing waterways datasets, this new global data provides an extension for those already working with data like TDX-Hydro. However, we also recognize a lag between resolution of water flow modeling and the resolution of waterways mapped by WaterNet. Recent efforts to serve historical and forecasted flood risk and water predictions globally, for example by the Group on Earth Observation Global Water Sustainability (GEOGLOWS), use a subset of the TDX-Hydro data due to this mismatch, as well as computational limitations \cite{Hales}. And so we expect there is likely catch up period needed to operationalize the vector data we create into existing scientific and analytical pipelines. At the same time, there is clear impetus to do this, for example, with recent work showing that ephemeral streams contribute up to 55\% of discharge exported from river systems in the United States \cite{Brinkerhoff}. There have also been recent exciting advances in flood prediction utilizing deep learning \cite{Nearing2024} which could aid this effort globally by reducing simulation costs. 

Obviously the specific downstream use case is relevant in how useful these new data are. In our previous work, we have found these new waterways maps to be extremely useful for capturing community infrastructure needs where other publicly available datasets fail \cite{Pierson2024_1}.  Scale of analysis and detection is critical here. To address localized and individual communities needs and to respond with localized anticipatory action to disasters to ensure communities have access to essential services, increased resolution, and capturing of additional waterways structures will become ever more important \cite{Lindersson20}.  As such we expect that this new approach and data may help in closing that important gap in cross-scale decision making, and aid humanitarian organizations in ways not possible with existing data. 

We do think that future research could use higher a resolution DEM, and even higher resolution satellite imagery, which we expect will improve the model outputs, and capture more fine waterways structures \cite{Moortgat2022}. At the same time, there are important trade-offs between scale, computation, acquisition, training and inference cost of features. What we present here is highly scalable, and could easily deployed or made operational with publicly available missions with modest GPU resources and storage requirements.  Further experiments may also improve on the representation of certain waterways in training and in the vectorization process. For example, we previously found that up-weighting swamps during training enabled the model to capture a catastrophic flooding event and humanitarian crisis in South Sudan \cite{Pierson2024_1, MSFsudan}.  Which model parametrization is most relevant will in turn depend on use case. Future efforts may blend, stack or ensemble different parametrization for different communities of users.

Critically, the model we distribute was only trained using data from the USA, where it learns to draw and reproduce waterways structures across a diverse range of hydrographic conditions with high structural accuracy.  In African countries where we have assessed community request data for trail bridges, independent point validation at the scale of predictions is also a good indication that a model built in the USA could be expanded to widely different geographies in a different continent \cite{Pierson2024_1}. While it does appear our model is able to generalize, and learn fundamental patterns from satellite data that transfer across widely different geographic and hydrographic contexts, continual efforts to collect more test data in new geographies, alongside inter-comparisons with future efforts like this, would be useful. 

It is our hope that the model and data presented here may help advance a number of applications - both scientific and humanitarian. They also hold value in fundamental discovery and characterization of the Earth system, of which water plays a critical role. How artificial intelligence and computer vision models can continue to assist in that discovery remains an exciting prospect for the future.

\section{Methods} \label{methods}

\subsection{Data overview}

\subsubsection{Data description}

Sentinel-2 Level-2A NRGB bands and Copernicus DEM GLO-30 data were acquired using the Microsoft Planetary Computer API (https://planetarycomputer.microsoft.com). To composite the Sentinel-2 data we obtained a list of all Sentinel-2 files for 2023 (and 2022 if required), and sorted that list by inverse of the proportion of missing and cloudy data, by running from most complete in the list $i$ to $i+3$, compositing scenes to replace data gaps and clouds with surface reflectances. We then applied the following transformation to each channel in the composite, ignoring any remaining masked out data: $$f(x) = \dfrac{255}{1+e^{-0.6x}}$$ and stored the composite as an unsigned 8-bit integer in EPSG:4326 (as a note, there is no guarantee that this normalization will keep channel ratios constant).

\subsubsection{Input data preparation}
The model has 10 channel inputs: Transformed Sentinel NRGB ($NRGB_t$), $NDVI$, $NDWI$, Shifted Elevation ($E_S$), Elevation x-delta ($\Delta_x E$), Elevation y-delta ($\Delta_y E$), elevation gradient ($\nabla E$).

These layers are obtained as follows:
\begin{enumerate}
    \item ($NRGB_t$) For this, we take first scale the NRGB values to 1, $$NRGB_{s} = NRGB/255,$$ we then transform the scaled data $$NRGB_t = 2*NRGB_s - 1.$$
    We keep the scaled NRGB data to use in our NDVI and NDWI computations.
    \item ($NDVI$) $$NDVI = \dfrac{N_s - R_s}{N_s + R_s}$$ Where $N_s$ and $R_s$ are the scaled $N$ and $R$ values.
    \item ($NDWI$) $$NDWI = \dfrac{G_s - N_s}{G_s + N_s}$$
    \item ($E_S$) For this we subtract the minimum elevation from each cell. $$E_S = E - \min(E).$$
    \item ($\Delta_{x} E$) For each cell at row $r$ and column $c$, we take 
    $$\Delta_{x} E[r, c] = (E[r, c+1] - E[r, c-1])/2$$
    \item ($\Delta_{y} E$) 
    $$\Delta_{y} E[r, c] = (E[r+1, c] - E[r-1, c])/2$$
    \item ($\nabla E$) $$\nabla E = (\Delta_{x} E^2 + \Delta_{y} E^2)^{0.5}$$
\end{enumerate}

\subsection{Training data preparation}
The National Hydrography Dataset (NHD) \cite{nhd} was utilized as training data, which is a high fidelity vector dataset of waterways in the United States. NHD data was burned to rasters that aligned with our Sentinel-2 data, with each fcode type (e.g. an identifier for each water type, such as rivers, streams, lakes, ditches, intermittent, ephemeral versions of each) assigned a different integer value. Using this identifier, we were able to give different waterway types different weights during the training process.

\subsection{Computer vision model}

 WaterNet is generally based on ideas from U-Net \cite{ronneberger2015unet} and ResNet \cite{he2015}. An overview of layers is provided in \ref{tab:model}, with expanded details on convolutions \ref{tab:convolution}, decoders \ref{tab:decoder}, encoders \ref{tab:encoder},  gated linear units \ref{tab:multiplication}, and residual layers \ref{tab:residual} - \ref{tab:residualB}. We maintain precision of raster outputs that are 20m globally.  

\subsection{Model training}

The model was trained on NHD data using augmentations such as flipping, rotating the original images, and dropping out 20\% of the input cells. A batch-size increase schedule was employed during training. The loss function was $$L(y, y_t) = 0.3\cdot\text{BCE}(y, y_t) + 0.7\cdot\text{TL}(y, y_t)$$ where BCE is Binary Cross-Entropy weighted by fcode type, and TL is Tanimoto loss \cite{fields} weighted by fcode type Supplementary Table  \ref{tab:edt2}. Weighting allowed us to adjust for the label imbalance and also to down weight or mask out fcode classes that we found created artifacts in the output (e.g. swamps, intermittent lakes, etc). 

Notably, the model was trained across a diversity of hydrographic conditions using labels from the National Hydrography Dataset  (e.g. with an identifier for each water type, such as rivers, streams, lakes, ditches, intermittent, ephemeral, called the fcode),  in two steps, starting  with a larger training set of smaller context ${\sim}$1.5M grids (244 x 244 pixels),  and followed with a ${\sim}$10x decrease in training samples but ${\sim}$ 10x increase in context, ${\sim}$90K grids (832 x 832 pixels). We have found this two step approach to be a useful for making location predictions across a diversity of contexts and water way types, while at the same time minimizing evaluation time and maximizing speed and alignment of waterways network structures in the final product.  The optimizer was stochastic gradient descent with momentum and L2 regularization. (lr=0.01, momentum=0.9, weight decay=0.0001). During training on both grid sizes we used a batch-size scheduler which would increase the batch size by twice the original batch size if the validation f1 score did not increase for 15 iterations.

\subsection{Vectorization process}

We vectorize the raster outputs of WaterNet to the TDX-Hydro backbone\cite{TDXHydro} . TDX-Hydro is a dataset developed by the National Geospatial-Intelligence Agency using the 12m TanDEM-X dataset. TDX-Hydro consists of waterways and their basins, one basin for each waterway in their dataset.  Our vectorization process involved several steps outlined below. All of the code used in the vectorization process is included with this paper.

\subsubsection{Connecting components}

We begin by connecting disconnected waterways components and clean model outputs (removing waterways that have some cells in the basin, but that should be considered in an adjacent basin). The steps are as follows:

\begin{enumerate}
    \item Cut the model output to the bounding box of a TDX-Hydro basin, buffered by 0.005 degrees, and burn the reference waterway corresponding to the basin to this raster. 
    \item Make a rounded copy of the model's output, and make a copy of the models output rescaled by $$f(x)=\min\left(1,\, \max\left(0,\, \dfrac{x-0.1}{0.5-0.1}\right)\right)$$
    This will be used to make weights in the graph, we ignore cells with a model probability less than $0.1$, and there is no additional penalty given to cells with a model output greater than $0.5$.
    \item Remove model waterways that intersect the basin, but that should connect to a different waterway in an adjacent basin.
        \begin{enumerate}
            \item  Make a grid representing each connected component in the rounded grid,  and use a connectivity rule to assign each connected region a distinct integer label to a grid, with  water being 8-connected (i.e. on the  horizontal, vertical, and diagonals), and land is 4-connected (i.e. only horizontal and vertical only).
            \item Remove a connected component if the cell with the minimum elevation falls outside of the basin and more than 50\% of the grid cells fall outside of the basin.
        \end{enumerate}
    \item Connect waterways to the reference waterway using a grid graph and least cost pathing
        \begin{enumerate}
            \item Cell midpoints are the nodes of the graph, and edges are added connecting adjacent cell nodes.
            \item Cells are only included if they have a nonzero scaled probability from the model's output, or touch the reference waterway.
            \item The edges are weighted. 
            \begin{enumerate}
                \item The weight of the edge from the source cell $(row_s, col_s)$ to the target $(row_t, col_t)$ is given by
\[
    weight = \begin{cases}
                 -\log_2(scaled_t)& \text{if } \Delta e <= 0 \\
                 \max(-\log_2(scaled_t) b* \Delta e, \Delta e)& \text{if } 0 < \Delta e
    \end{cases}
\]
                where $$\Delta e = elevation_t - elevation_s$$ and $scaled_t$ is the scaled probability value from (2).
            \end{enumerate}
            \item We iteratively run the least cost path algorithm starting at minimum elevation cells for the disconnected waterways, allowing the algorithm to search further with each iteration, and including the newly connected waterways in each additional search.    
        \end{enumerate}
\end{enumerate}

\subsubsection{Thinning}
Next we run a thinning algorithm on the connected data. The idea is to remove all cells from the model outputs that wont change the topology of waterways in the basin (i.e. that won't change the number of connected components), leaving only the center most cells. The algorithm is canonical. The intuition is that if we have two adjacent rows of cells labeled as waterways, then we want to thin (remove) the cells with higher elevation.

Cells are labeled as either skeleton, interior, or (potentially) removable (a point that can be removed without altering the topology). Defined as:

\begin{enumerate}
    \item A cell is labeled a skeleton cell if it is touching at most one other waterway cell, or if its removal would change the connectedness of the waterway. That is, if its removal would turn a single waterway into two or more waterways which were no longer connected.
    \item A cell is labeled an interior cell if its removal would introduce a hole in the waterway.
    \item A cell is labeled a (potentially) removable cell if it is neither a skeleton cell nor an interior cell. It is potentially removable because as other cells are removed, a removable cell may become a skeleton cell.
\end{enumerate}

As a note, in this process all cells intersecting a TDX-Hydro stream are labeled as skeleton.

We then run an algorithm with the following pseudocode:

\begin{verbatim} 
    while length(removable_cells) > 0:
        new_removable_cells = []
        sort removable_cells by descending elevation.
        for removable_cell in removable_cells:
            if still removable:
                remove cell
            else:
                add cell to skeleton_cells
            for interior_cells adjacent to removable_cell:
                if interior_cell is removable:
                    add interior_cell to new_removable_cells
        removable_cells = new_removable_cells
\end{verbatim}

\subsubsection{Vectorization}
We then vectorize the thinned grid in a two step process.  First we connect the thinned model outputs to each other by using the midpoints of the cells as nodes and we connect all adjacent cells. During this process we keep track if each waterway segment (ie a waterway that only intersects other waterways at its head and tail). Next we connect each waterway segment to the TDX waterways, connecting each generated segment from exactly one node.

\subsubsection{Removing cycles}
We remove cycles (loops) by computing the least cost path from each of the model's nodes to their intersection points with the TDX waterways. The weights in the least cost path are given by $$f(x) = \max(0, \Delta \text{elevation}).$$
We keep every edge that appears in one of the least cost paths.

\subsubsection{Adding stream order}
As a final step, we add the Strahler stream order to the waterways, we give each waterway segment a unique ID, and for each segment we note the ID of any source waterway, and the unique target ID for each waterway, using $-1$ in place of any missing data. When computing the Strahler stream order for the TDX-Hydro waterways, we compare to the Strahler stream order in the TDX-Hydro dataset, and take the max of the new order and the old order.

\subsection{Waterway type analysis}

To compute waterway type we labeled every point in our dataset with the fcode description of the nearest waterway (shortest Euclidean distance using latitude and longitude)  in the NHD dataset within a maximum distance of 0.001 degrees. Points that didn't fall within 0.001 degrees of an NHD waterway were labeled as 'Unknown'.  In total, we found $78.05\%$ ($356,118,769/456,295,388$) of our waterway points had a known label, i.e they fell within $0.001$ degrees of an item in the NHD dataset.  $59.70\%$ ($272,415,638/456,295,388$) of all points, $76.50\%$ ($272,415,638/356,118,769$) of the known labeled points, had a label of Stream/River: Perennial, Stream/River: Intermittent, or Stream/River: Ephemeral. Of the $272,415,638$ points with those labels, $17.34\%$ were labeled perennial, $59.86\%$ were labeled intermittent, and $22.80\%$ were labeled ephemeral.

\printbibliography{}

@article{Brinkerhoff,
author = {Craig B. Brinkerhoff  and Colin J. Gleason  and Matthew J. Kotchen  and Douglas A. Kysar  and Peter A. Raymond },
title = {Ephemeral stream water contributions to United States drainage networks},
journal = {Science},
volume = {384},
number = {6703},
pages = {1476-1482},
year = {2024},
doi = {10.1126/science.adg9430},
}

@data{YY2XMG_2024,
author = {Pierson, Matthew and Mehrabi, Zia},
publisher = {Harvard Dataverse},
title = {{WaterNet Outputs and Code}},
year = {2024},
version = {V1},
doi = {10.7910/DVN/YY2XMG},
url = {https://doi.org/10.7910/DVN/YY2XMG}
}

@article{Nearing2024,
  title = {Global prediction of extreme floods in ungauged watersheds},
  volume = {627},
  ISSN = {1476-4687},
  url = {http://dx.doi.org/10.1038/s41586-024-07145-1},
  DOI = {10.1038/s41586-024-07145-1},
  number = {8004},
  journal = {Nature},
  publisher = {Springer Science and Business Media LLC},
  author = {Nearing,  Grey and Cohen,  Deborah and Dube,  Vusumuzi and Gauch,  Martin and Gilon,  Oren and Harrigan,  Shaun and Hassidim,  Avinatan and Klotz,  Daniel and Kratzert,  Frederik and Metzger,  Asher and Nevo,  Sella and Pappenberger,  Florian and Prudhomme,  Christel and Shalev,  Guy and Shenzis,  Shlomo and Tekalign,  Tadele Yednkachw and Weitzner,  Dana and Matias,  Yossi},
  year = {2024},
  month = mar,
  pages = {559–563}
}

@article{Hales,
author = {Hales, Riley C. and Nelson, E. James and Souffront, Michael and Gutierrez, Angelica L. and Prudhomme, Christel and Kopp, Steve and Ames, Daniel P. and Williams, Gustavious P. and Jones, Norman L.},
title = {Advancing global hydrologic modeling with the GEOGloWS ECMWF streamflow service},
journal = {Journal of Flood Risk Management},
volume = {n/a},
number = {n/a},
pages = {e12859},
doi = {https://doi.org/10.1111/jfr3.12859}
}

@misc{ronneberger2015unet,
      title={U-Net: Convolutional Networks for Biomedical Image Segmentation}, 
      author={Olaf Ronneberger and Philipp Fischer and Thomas Brox},
      year={2015},
      eprint={1505.04597},
      archivePrefix={arXiv},
      primaryClass={cs.CV}
}

@misc{he2015,
      title={Deep Residual Learning for Image Recognition}, 
      author={Kaiming He and Xiangyu Zhang and Shaoqing Ren and Jian Sun},
      year={2015},
      eprint={1512.03385},
      archivePrefix={arXiv},
      primaryClass={cs.CV},
      url={https://arxiv.org/abs/1512.03385}, 
}

@article{Lehner,
author = {Lehner, Bernhard and Grill, Gunther},
title = {Global river hydrography and network routing: baseline data and new approaches to study the world's large river systems},
journal = {Hydrological Processes},
volume = {27},
number = {15},
pages = {2171-2186},
doi = {https://doi.org/10.1002/hyp.9740},
year = {2013}
}

@article{Lindersson20,
author = {Lindersson, Sara and Brandimarte, Luigia and Mård, Johanna and Di Baldassarre, Giuliano},
title = {A review of freely accessible global datasets for the study of floods, droughts and their interactions with human societies},
journal = {WIREs Water},
volume = {7},
number = {3},
pages = {e1424},
doi = {https://doi.org/10.1002/wat2.1424},
year = {2020}
}

@article{
Allen2018,
author = {George H. Allen  and Tamlin M. Pavelsky },
title = {Global extent of rivers and streams},
journal = {Science},
volume = {361},
number = {6402},
pages = {585-588},
year = {2018},
doi = {10.1126/science.aat0636}
}

@ARTICLE{Messager2021-yp,
  title    = "Global prevalence of non-perennial rivers and streams",
  author   = "Messager, Mathis Loic and Lehner, Bernhard and Cockburn,
              Charlotte and Lamouroux, Nicolas and Pella, Herv{\'e} and
              Snelder, Ton and Tockner, Klement and Trautmann, Tim and Watt,
              Caitlin and Datry, Thibault",
  journal  = "Nature",
  volume   =  594,
  number   =  7863,
  pages    = "391--397",
  month    =  jun,
  year     =  2021
}

@misc{MSFsudan,
title= {Catastrophic floods cause mass displacement and humanitarian crisis},
    url= {https://www.msf.org/catastrophic-floods-cause-mass-displacement-and-escalate-humanitarian-crisis-south-sudan},
   year = {2022},
author= {Medecins Sans Frontieres}
}

@article{Moortgat2022,
title = {Deep learning models for river classification at sub-meter resolutions from multispectral and panchromatic commercial satellite imagery},
journal = {Remote Sensing of Environment},
volume = {282},
pages = {113279},
year = {2022},
issn = {0034-4257},
doi = {https://doi.org/10.1016/j.rse.2022.113279},
author = {Joachim Moortgat and Ziwei Li and Michael Durand and Ian Howat and Bidhyananda Yadav and Chunli Dai},
}

@misc{nhd,
  title     = "National Hydrography Dataset ({NHD})",
  publisher = "US Geological Survey",
  howpublished = "\url{https://apps.nationalmap.gov/downloader/}",
  year      =  2023
}

@misc{OpenStreetMap,
   author = {{OpenStreetMap contributors}},
   title = {{Data retrieved from https://download.geofabrik.de/}},
   howpublished = "\url{ https://www.openstreetmap.org }",
   year=2023,
 }

@ARTICLE{Pekel2016-bt,
  title    = "High-resolution mapping of global surface water and its long-term
              changes",
  author   = "Pekel, Jean-Fran{\c c}ois and Cottam, Andrew and Gorelick, Noel
              and Belward, Alan S",
  journal  = "Nature",
  volume   =  540,
  number   =  7633,
  pages    = "418--422",
  month    =  dec,
  year     =  2016
}

@misc{Pierson2024_1,
      title={Deep learning waterways for rural infrastructure development}, 
      author={Matthew Pierson and Zia Mehrabi},
      year={2024},
      eprint={1.1},
        doi={ 	
https://doi.org/10.48550/arXiv.2411.13590},
      archivePrefix={arXiv},
      primaryClass={cs.CV}
}

@article{carlson_2024_tdxhydro,
  author = {Carlson, Kimberly A and Levin, Heather K and Morris, Amy L and Candela, Salvatore G and Morales, M and Huening, Vincent G and Fredericks, Jaimeson G},
  month = {05},
  title = {TDX-Hydro: Global High-Resolution Hydrography Derived from TanDEM-X},
  doi = {10.22541/essoar.171629686.65893579/v1},
  year = {2024},
  journal = {Authorea (Authorea)}
}

@misc{TDXHydro,
   author = {{The National Geospatial-Intelligence Agency}},
   title = {TDX-Hydro},
   howpublished = "\url{https://earth-info.nga.mil/}",
   year= 2024
 }

@misc{Microsoftroads2021,
   author = {{Microsoft}},
   title = {Road Detections},
year      =  2021,
   howpublished = "\url{https://github.com/microsoft/RoadDetections}",

 }

\section{Acknowledgments}
This project was partially funded by Bridges to Prosperity under the grant “Remote Impact Assessment of Rural Infrastructure Development” to the Better Planet Laboratory (https://betterplanetlab.com/). The authors would like to thank Abbie Noriega, Kyle Shirley, Cam Kruse for feedback.

\section{Author Contributions}
MP designed and implemented WaterNet with input from ZM. MP and ZM interpreted the results. MP and ZM wrote the paper. 

\section{Competing Interests}
None

\section{Figures and Tables}

\begin{landscape}
\begin{figure}
    \centering
    \includegraphics[width=10in]{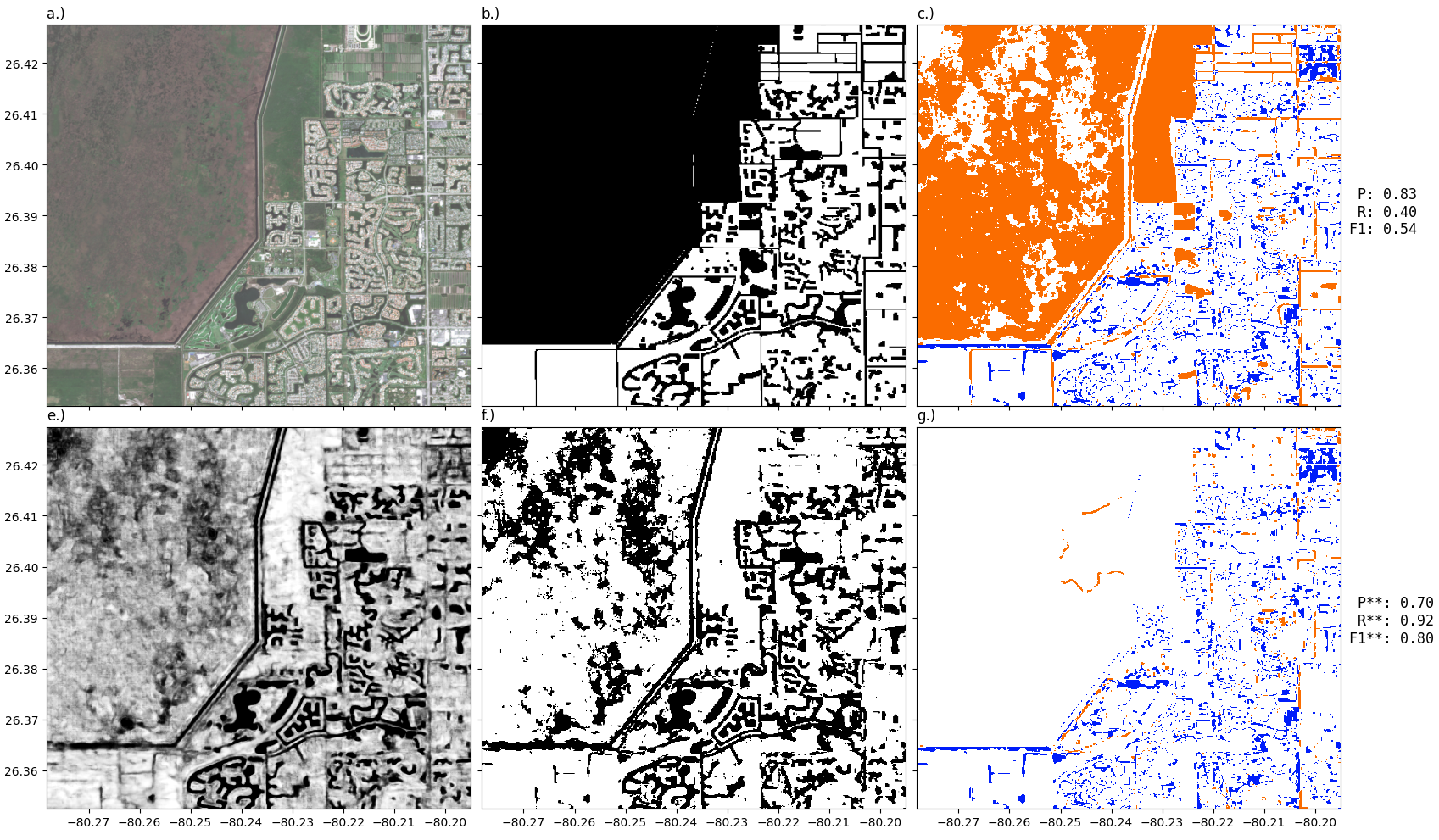}
    \caption { \textbf{Impact of masking waterway types on performance} a) Input Sentinel-2 data, b) Burned National Hydrography Data (NHD), e) Model output, f) Model output rounded at a probability of 0.5, c) Difference between the rounded output and the NHD data, g) The same difference as in c) but ignoring water(way) types that were masked out during training (swamps in this figure). WaterNet is tuned to detect rivers and streams and these are the primary target of our training, but we evaluate the model performance on all waterway types.Blue = false positive, Orange = false negative}
    \label{fig:mask}
\end{figure}

\begin{table}
    \centering
\begin{NiceTabular}{|c||c|c|c||c|c|c||c|c|c||c|}
	\hline
	\textbf{\makecell{Subset of Images}} & \textbf{\makecell{P}} & \textbf{\makecell{R}} & \textbf{\makecell{F1}} & \textbf{\makecell{P*}} & \textbf{\makecell{R*}} & \textbf{\makecell{F1*}} & \textbf{\makecell{P**}} & \textbf{\makecell{R**}} & \textbf{\makecell{F1**}} & \textbf{\makecell{Data Percent}} \\
	\hline
    \hline
	\textbf{\makecell{All Data\\Included}} & 0.7200 & 0.6034 & 0.6566 & 0.8235 & 0.6446 & 0.7232 & 0.6888 & 0.7236 & 0.7058 & 100.00\% \\
	\hline
	\textbf{\makecell{Data With Any\\Mask Type Excluded}} & 0.7384 & 0.7665 & 0.7522 & 0.8481 & 0.8271 & 0.8375 & 0.7384 & 0.7665 & 0.7522 & 22.71\% \\
	\hline
	\textbf{\makecell{Canals\\ Excluded}} & 0.6953 & 0.6554 & 0.6748 & 0.8144 & 0.7112 & 0.7593 & 0.6726 & 0.7098 & 0.6907 & 67.34\% \\
	\hline
	\textbf{\makecell{Canals\\Included}} & 0.7653 & 0.5331 & 0.6284 & 0.8392 & 0.5577 & 0.6701 & 0.7216 & 0.7511 & 0.7360 & 32.66\% \\
	\hline
	\textbf{\makecell{Intermittent Lakes\\Excluded}} & 0.7941 & 0.6305 & 0.7029 & 0.8726 & 0.6593 & 0.7511 & 0.7603 & 0.7818 & 0.7709 & 46.22\% \\
	\hline
	\textbf{\makecell{Intermittent Lakes\\Included}} & 0.6219 & 0.5627 & 0.5908 & 0.7521 & 0.6211 & 0.6803 & 0.6035 & 0.6508 & 0.6263 & 53.78\% \\
	\hline
	\textbf{\makecell{Swamps\\Excluded}} & 0.6670 & 0.6920 & 0.6793 & 0.7872 & 0.7607 & 0.7737 & 0.6598 & 0.7156 & 0.6866 & 75.60\% \\
	\hline
	\textbf{\makecell{Swamps\\Included}} & 0.8257 & 0.5004 & 0.6231 & 0.8897 & 0.5175 & 0.6544 & 0.7651 & 0.7425 & 0.7536 & 24.40\% \\
	\hline
\end{NiceTabular}
    \caption{\textbf{Model Performance}. The model's precision (P), recall (R) and F1 scores using different subsets of the test data. For example, 'Data With Canals Excluded' means that we excluded any of the n= 10,887 test data that had a canal anywhere in the image, and 'Data With Canals Included' means that a canal must appear in the image to be in the test set. Data Percent = percentage of n test data used in the calculation. P, R, and F1 indicate standard pixel scores. P*, R*, and F1* are the scores if we ignore errors that are adjacent to a correct true and a correct false prediction. This ignores errors due to the model's prediction being too thick, an example can be seen in Figure \ref{fig:thick}. P**, R**, and F1** are the scores when we mask out data types that were masked during training (swamps, canals, ditches, drainage, intermittent lakes, playas), an example can be seen in Figure \ref{fig:mask}.}
\label{tab:scores}
\end{table}

\begin{figure}
    \centering
    \includegraphics[width=7in]{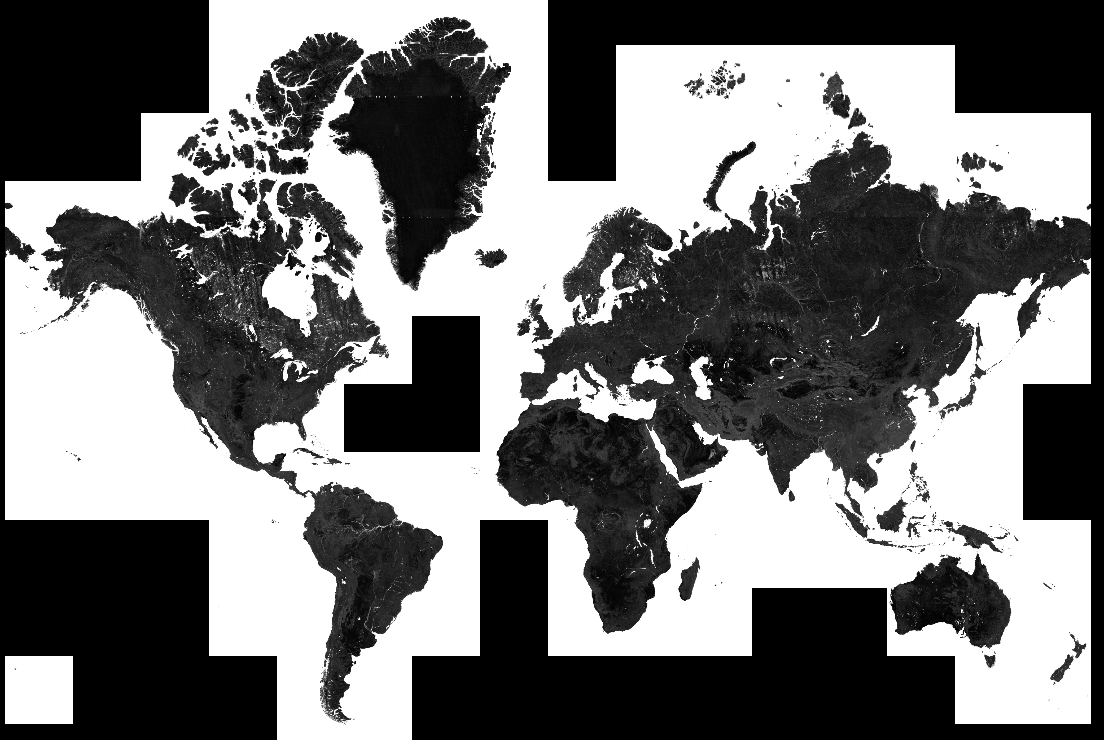}
    \caption{\textbf{Global extent of WaterNet predictions the year circa 2023}. A  raster output of the global extent at 20m x 20m resolution, predicted from 10 satellite features derived from cloud-free mosaic of 10m Sentinel-2 Level-2A NRGB bands and the 30m Copernicus Digital Surface Model. A waterway probability of >=0.5 is shown in white, with all other land shown in black. Note some areas of the ocean are masked.}
    \label{fig:global}
\end{figure}
\end{landscape}

\newpage

\begin{figure}[H]
    \centering
    \includegraphics[width=6in]{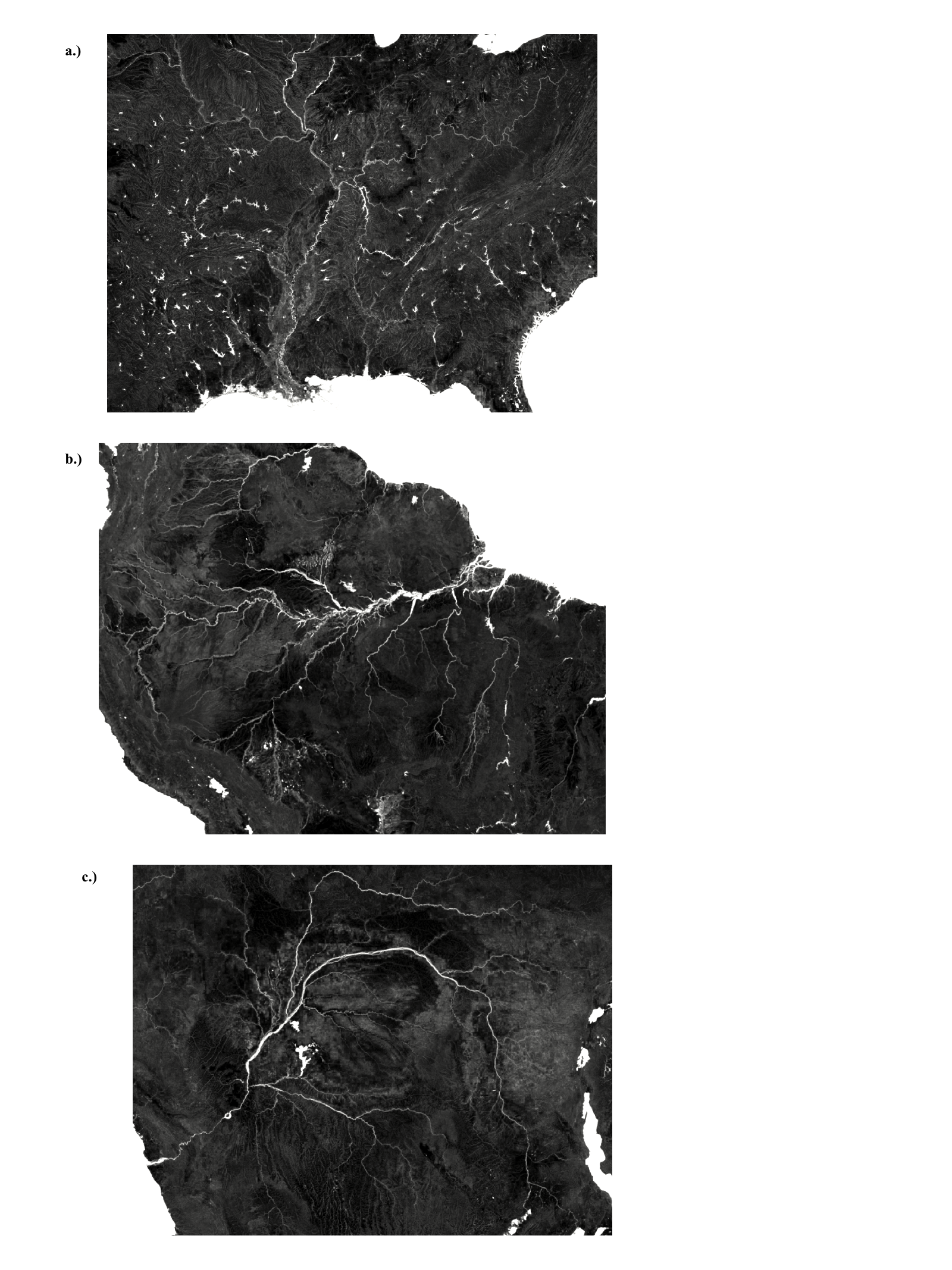}
     \caption{\textbf{Regional examples of WaterNet predictions the year circa 2023}. Major river systems are shown including, a) Mississippi system in United States of America b) Amazon system in South America and c) the Congo river system in Central Africa. A waterway probability of >=0.5 is shown in white, with all other land shown in black.}
    \label{fig:regional}
\end{figure}

\renewcommand{\arraystretch}{1.25}
\begin{table}
    \centering
    \begin{NiceTabular}{|c|c|c|c|}
    \hline
    \textbf{\makecell{Data Source}} & \textbf{\makecell{Stream Order}} & \textbf{\makecell{Total Length in\\kilometers}} & \textbf{Status} \\
    \hline
    \hline
    WaterNet & 1 & 75,139,170 & New \\
    \hline
    WaterNet & 2 & 38,260,579 & New \\
    \hline
    WaterNet & 3 & 10,470,285 & New \\
    \hline
    WaterNet & 4 & 766,109 & New \\
    \hline
    WaterNet & 5 & 41,599 & New \\
    \hline
    WaterNet & 6 & 579 & New \\
    \hline
    TDX-Hydro & 1 & 5,544,968 & Existing \\
    \hline
    TDX-Hydro & 2 & 6,885,653 & Existing \\
    \hline
    TDX-Hydro & 3 & 15,045,456 & Existing \\
    \hline
    TDX-Hydro & 4 & 14,410,753 & Existing \\
    \hline
    TDX-Hydro & 5 & 6,785,578 & Existing \\
    \hline
    TDX-Hydro & 6 & 3,332,659 & Existing \\
    \hline
    TDX-Hydro & 7 & 1,667,972 & Existing \\
    \hline
    TDX-Hydro & 8 & 787,313 & Existing \\
    \hline
    TDX-Hydro & 9 & 337,797 & Existing \\
    \hline
    TDX-Hydro & 10 & 115,965 & Existing \\
    \hline
    TDX-Hydro & 11 & 31,154 & Existing \\
    \hline
    TDX-Hydro & 12 & 4,999 & Existing \\
    \hline
\end{NiceTabular}

    \caption{\textbf{New global waterways mapped}. All WaterNet stream lengths shown are in addition to the existing stream lengths shown for TDX-Hydro, which represents the prior current state of knowledge. In total, WaterNet adds nearly 125 million kilometers of waterway to the 55 million kilometers in the TDX-Hydro dataset.  The significant gains arise from order 1, 2 and 3 streams, representing more than 75M, 38M and 10M new waterways mapped, respectively.  Waterways that intersect lakes are removed from these calculations. Many of these new mapped waterways are likely to be intermittent and ephemeral, overlooked waterways that can be important for people and nature.}
    \label{tab:lenadded}
\end{table}

\section{Data Availability}
Global raster and vector outputs are available at the Harvard Dataverse doi: 10.7910/DVN/YY2XMG under a CC-BY-SA 4.0 license.

\section{Code Availability}
WaterNet code is freely available at the Harvard Dataverse doi: 10.7910/DVN/YY2XMG under a GNU GPL v3 license.

\newpage

\section{Supplementary Information}

\begin{table}[h]
\setcounter{table}{0}
	\tiny
    \centering
\begin{NiceTabular}{|c||c|c|c||c|c|c||c|c|c||c|}
	\hline
	\textbf{\makecell{Data Type}} & \textbf{\makecell{P}} & \textbf{\makecell{R}} & \textbf{\makecell{F1}} & \textbf{\makecell{P*}} & \textbf{\makecell{R*}} & \textbf{\makecell{F1*}} & \textbf{\makecell{P**}} & \textbf{\makecell{R**}} & \textbf{\makecell{F1**}} & \textbf{\makecell{Data Percent}} \\
	\hline
    \hline
	\textbf{\makecell{Using \\ All Data}} & 0.7200 & 0.6034 & 0.6566 & 0.8235 & 0.6446 & 0.7232 & 0.6888 & 0.7236 & 0.7058 & 100.0\% \\
	\hline
	\textbf{\makecell{Using \\ HU4 103 }} & 0.7460 & 0.7494 & 0.7477 & 0.8178 & 0.7853 & 0.8012 & 0.7460 & 0.7494 & 0.7477 & 2.45\% \\
	\hline
	\textbf{\makecell{Using \\ HU4 204 }} & 0.7760 & 0.8045 & 0.7900 & 0.8343 & 0.8420 & 0.8382 & 0.7593 & 0.8511 & 0.8026 & 6.34\% \\
	\hline
	\textbf{\makecell{Using \\ HU4 309 }} & 0.9361 & 0.4736 & 0.6290 & 0.9536 & 0.4784 & 0.6371 & 0.8727 & 0.9100 & 0.8910 & 5.86\% \\
	\hline
	\textbf{\makecell{Using \\ HU4 403 }} & 0.9059 & 0.6809 & 0.7774 & 0.9393 & 0.6959 & 0.7995 & 0.9053 & 0.8803 & 0.8926 & 8.08\% \\
	\hline
	\textbf{\makecell{Using \\ HU4 505 }} & 0.5254 & 0.7160 & 0.6060 & 0.6972 & 0.8462 & 0.7645 & 0.5243 & 0.7287 & 0.6098 & 4.90\% \\
	\hline
	\textbf{\makecell{Using \\ HU4 601 }} & 0.6181 & 0.6305 & 0.6242 & 0.7903 & 0.7225 & 0.7549 & 0.6180 & 0.6308 & 0.6243 & 6.40\% \\
	\hline
	\textbf{\makecell{Using \\ HU4 701 }} & 0.8314 & 0.8281 & 0.8297 & 0.8809 & 0.8580 & 0.8693 & 0.8295 & 0.8640 & 0.8464 & 7.52\% \\
	\hline
	\textbf{\makecell{Using \\ HU4 805 }} & 0.7522 & 0.5366 & 0.6264 & 0.8314 & 0.5761 & 0.6806 & 0.7368 & 0.6399 & 0.6849 & 2.33\% \\
	\hline
	\textbf{\makecell{Using \\ HU4 904 }} & 0.6921 & 0.5733 & 0.6271 & 0.8240 & 0.6262 & 0.7116 & 0.6875 & 0.5893 & 0.6346 & 0.43\% \\
	\hline
	\textbf{\makecell{Using \\ HU4 1008 }} & 0.5749 & 0.6511 & 0.6106 & 0.7244 & 0.7383 & 0.7313 & 0.5695 & 0.6849 & 0.6219 & 8.42\% \\
	\hline
	\textbf{\makecell{Using \\ HU4 1110 }} & 0.5721 & 0.6712 & 0.6177 & 0.7155 & 0.7471 & 0.7310 & 0.5650 & 0.7321 & 0.6378 & 7.09\% \\
	\hline
	\textbf{\makecell{Using \\ HU4 1203 }} & 0.6539 & 0.6470 & 0.6504 & 0.7713 & 0.7151 & 0.7421 & 0.6434 & 0.7271 & 0.6827 & 7.04\% \\
	\hline
	\textbf{\makecell{Using \\ HU4 1302 }} & 0.5546 & 0.5594 & 0.5570 & 0.7099 & 0.6439 & 0.6753 & 0.5444 & 0.5965 & 0.5693 & 10.13\% \\
	\hline
	\textbf{\makecell{Using \\ HU4 1403 }} & 0.5447 & 0.6617 & 0.5975 & 0.6959 & 0.7552 & 0.7243 & 0.5423 & 0.6794 & 0.6031 & 3.21\% \\
	\hline
	\textbf{\makecell{Using \\ HU4 1505 }} & 0.5626 & 0.5246 & 0.5429 & 0.6992 & 0.6000 & 0.6458 & 0.5559 & 0.5658 & 0.5608 & 6.27\% \\
	\hline
	\textbf{\makecell{Using \\ HU4 1603 }} & 0.5430 & 0.6457 & 0.5899 & 0.6454 & 0.7083 & 0.6754 & 0.4628 & 0.6411 & 0.5375 & 6.20\% \\
	\hline
	\textbf{\makecell{Using \\ HU4 1708 }} & 0.7064 & 0.4948 & 0.5820 & 0.8479 & 0.5415 & 0.6609 & 0.7004 & 0.5015 & 0.5844 & 2.05\% \\
	\hline
	\textbf{\makecell{Using \\ HU4 1804 }} & 0.6832 & 0.4060 & 0.5094 & 0.8493 & 0.4510 & 0.5891 & 0.6667 & 0.4445 & 0.5334 & 5.48\% \\
	\hline
\end{NiceTabular}
    \small
     \def\tablename{Supplementary Table}
    \caption{Test statistics for the individual HU4 test regions.}
    \label{tab:hu4}
\end{table}

\begin{table}
\renewcommand{\arraystretch}{1.5}
\begin{multicols}{3}
\tiny

    \centering
\begin{NiceTabular}{|c|c|c|}
	\hline
	\textbf{\makecell{Basin ID}} & \textbf{\makecell{Data\\Source}} & \textbf{\makecell{Total\\Length in\\kilometers}} \\
	\hline
	\hline
    1020000010 & TDX-Hydro & 1,430,476 \\
	1020000010 & WaterNet & 2,929,845 \\
	\hline
	1020011530 & TDX-Hydro & 2,087,073 \\
	1020011530 & WaterNet & 2,867,418 \\
	\hline
	1020018110 & TDX-Hydro & 1,890,022 \\
	1020018110 & WaterNet & 3,913,021 \\
	\hline
	1020021940 & TDX-Hydro & 1,572,211 \\
	1020021940 & WaterNet & 3,061,513 \\
	\hline
	1020027430 & TDX-Hydro & 3,207,612 \\
	1020027430 & WaterNet & 4,556,487 \\
	\hline
	1020034170 & TDX-Hydro & 1,301,042 \\
	1020034170 & WaterNet & 2,499,352 \\
	\hline
	1020035180 & TDX-Hydro & 249,458 \\
	1020035180 & WaterNet & 909,550 \\
	\hline
	1020040190 & TDX-Hydro & 1,107,758 \\
	1020040190 & WaterNet & 1,119,731 \\
	\hline
	2020000010 & TDX-Hydro & 543,499 \\
	2020000010 & WaterNet & 1,838,425 \\
	\hline
	2020003440 & TDX-Hydro & 1,048,125 \\
	2020003440 & WaterNet & 2,351,880 \\
	\hline
	2020018240 & TDX-Hydro & 626,194 \\
	2020018240 & WaterNet & 1,692,455 \\
	\hline
	2020024230 & TDX-Hydro & 588,562 \\
	2020024230 & WaterNet & 946,614 \\
	\hline
	2020033490 & TDX-Hydro & 175,025 \\
	2020033490 & WaterNet & 488,531 \\
	\hline
	2020041390 & TDX-Hydro & 424,915 \\
	2020041390 & WaterNet & 749,045 \\
	\hline
	2020057170 & TDX-Hydro & 39,654 \\
	2020057170 & WaterNet & 79,330 \\
	\hline
	2020065840 & TDX-Hydro & 1,456,936 \\
	2020065840 & WaterNet & 2,742,739 \\
	\hline
	2020071190 & TDX-Hydro & 2,698,025 \\
	2020071190 & WaterNet & 7,306,654 \\
	\hline
	3020000010 & TDX-Hydro & 1,496,899 \\
	3020000010 & WaterNet & 2,444,513 \\
	\hline
	3020003790 & TDX-Hydro & 972,008 \\
	3020003790 & WaterNet & 2,278,033 \\
	\hline
	3020005240 & TDX-Hydro & 428,410 \\
	3020005240 & WaterNet & 734,927 \\
	\hline
	3020008670 & TDX-Hydro & 932,256 \\
	3020008670 & WaterNet & 1,888,061 \\
	\hline
	3020009320 & TDX-Hydro & 1,062,511 \\
	3020009320 & WaterNet & 2,777,580 \\
	\hline
	3020024310 & TDX-Hydro & 143,345 \\
	3020024310 & WaterNet & 353,626 \\
	\hline
\end{NiceTabular}
 \def\tablename{Supplementary Table}

\columnbreak

\begin{NiceTabular}{|c|c|c|}
	\hline
	\textbf{\makecell{Basin ID}} & \textbf{\makecell{Data\\Source}} & \textbf{\makecell{Total\\Length in\\kilometers}} \\
	\hline
    \hline
  	4020000010 & TDX-Hydro & 1,298,302 \\
	4020000010 & WaterNet & 3,143,166 \\
	\hline
	4020006940 & TDX-Hydro & 1,827,603 \\
	4020006940 & WaterNet & 8,309,523 \\
	\hline
	4020015090 & TDX-Hydro & 929,593 \\
	4020015090 & WaterNet & 3,590,183 \\
	\hline
	4020024190 & TDX-Hydro & 1,988,734 \\
	4020024190 & WaterNet & 5,816,401 \\
	\hline
	4020034510 & TDX-Hydro & 186,527 \\
	4020034510 & WaterNet & 762,913 \\
	\hline
	4020050210 & TDX-Hydro & 823,313 \\
	4020050210 & WaterNet & 2,062,830 \\
	\hline
	4020050220 & TDX-Hydro & 1,124,455 \\
	4020050220 & WaterNet & 2,059,992 \\
	\hline
	4020050290 & TDX-Hydro & 789,603 \\
	4020050290 & WaterNet & 1,916,223 \\
	\hline
	4020050470 & TDX-Hydro & 229,214 \\
	4020050470 & WaterNet & 740,938 \\
 	\hline
 	5020000010 & TDX-Hydro & 305,011 \\
	5020000010 & WaterNet & 943,362 \\
	\hline
	5020015660 & TDX-Hydro & 511,422 \\
	5020015660 & WaterNet & 1,639,450 \\
	\hline
	5020037270 & TDX-Hydro & 401,052 \\
	5020037270 & WaterNet & 1,069,681 \\
	\hline
	5020049720 & TDX-Hydro & 2,475,481 \\
	5020049720 & WaterNet & 3,010,163 \\
	\hline
	5020054880 & TDX-Hydro & 405 \\
	5020054880 & WaterNet & 1,827 \\
	\hline
	5020055870 & TDX-Hydro & 17,215 \\
	5020055870 & WaterNet & 72,051 \\
	\hline
 	5020082270 & TDX-Hydro & 118,310 \\
	5020082270 & WaterNet & 407,088 \\
	\hline
	6020000010 & TDX-Hydro & 894,029 \\
	6020000010 & WaterNet & 2,628,113 \\
	\hline
	6020006540 & TDX-Hydro & 2,812,076 \\
	6020006540 & WaterNet & 6,595,560 \\
	\hline
	6020008320 & TDX-Hydro & 786,710 \\
	6020008320 & WaterNet & 2,316,011 \\
	\hline
	6020014330 & TDX-Hydro & 1,578,636 \\
	6020014330 & WaterNet & 2,892,433 \\
	\hline
	6020017370 & TDX-Hydro & 717,782 \\
	6020017370 & WaterNet & 1,072,130 \\
	\hline
	6020021870 & TDX-Hydro & 398,774 \\
	6020021870 & WaterNet & 1,447,621 \\
	\hline
	6020029280 & TDX-Hydro & 203,900 \\
	6020029280 & WaterNet & 791,703 \\
    \hline
\end{NiceTabular}
 \def\tablename{Supplementary Table}

\columnbreak

\centering
\begin{NiceTabular}{|c|c|c|}
	\hline
	\textbf{\makecell{Basin ID}} & \textbf{\makecell{Data\\Source}} & \textbf{\makecell{Total\\Length in\\kilometers}} \\
    \hline
    \hline
	7020000010 & TDX-Hydro & 1,127,584 \\
	7020000010 & WaterNet & 4,211,940 \\
	\hline
	7020014250 & TDX-Hydro & 507,823 \\
	7020014250 & WaterNet & 1,634,549 \\
	\hline
	7020021430 & TDX-Hydro & 707,232 \\
	7020021430 & WaterNet & 918,577 \\
	\hline
	7020024600 & TDX-Hydro & 637,190 \\
	7020024600 & WaterNet & 1,665,120 \\
	\hline
	7020038340 & TDX-Hydro & 447,657 \\
	7020038340 & WaterNet & 987,950 \\
	\hline
	7020046750 & TDX-Hydro & 1,516,722 \\
	7020046750 & WaterNet & 3,639,502 \\
	\hline
	7020047840 & TDX-Hydro & 980,613 \\
	7020047840 & WaterNet & 2,639,528 \\
	\hline
	7020065090 & TDX-Hydro & 93,888 \\
	7020065090 & WaterNet & 313,134 \\
	\hline
	8020000010 & TDX-Hydro & 683,900 \\
	8020000010 & WaterNet & 1,868,562 \\
	\hline
	8020008900 & TDX-Hydro & 595,843 \\
	8020008900 & WaterNet & 1,297,539 \\
	\hline
	8020010700 & TDX-Hydro & 143,121 \\
	8020010700 & WaterNet & 297,293 \\
	\hline
	8020020760 & TDX-Hydro & 44,731 \\
	8020020760 & WaterNet & 80,689 \\
	\hline
	8020022890 & TDX-Hydro & 81,504 \\
	8020022890 & WaterNet & 172,068 \\
	\hline
	8020032840 & TDX-Hydro & 128,190 \\
	8020032840 & WaterNet & 328,304 \\
	\hline
	8020044560 & TDX-Hydro & 176,057 \\
	8020044560 & WaterNet & 353,592 \\
	\hline
	9020000010 & TDX-Hydro & 1,178,049 \\
	9020000010 & WaterNet & 451,282 \\
	\hline
\end{NiceTabular}
\end{multicols}
 \def\tablename{Supplementary Table}
    \caption{The total length of waterways in each Hydrobasins level 2 basin by data source, excluding waterways that intersect polygons in the HydroLakes dataset.}
    \label{tab:lenbybasin}
\end{table}

\renewcommand{\arraystretch}{1.}

\newpage
\clearpage

\begin{table}[H]
\tiny
    \centering
\begin{NiceTabular}{|c|c||c|c|c|c|c|c|c|c|c|c|}
	\hline
	\textbf{\makecell{Stream\\Order}} & \textbf{\makecell{Hydrographic\\Category}}& \textbf{\makecell{All\\Regions}} & \textbf{\makecell{HU2\\01}} & \textbf{\makecell{HU2\\02}} & \textbf{\makecell{HU2\\03}} & \textbf{\makecell{HU2\\04}} & \textbf{\makecell{HU2\\05}} & \textbf{\makecell{HU2\\06}} & \textbf{\makecell{HU2\\07}} & \textbf{\makecell{HU2\\08}} & \textbf{\makecell{HU2\\09}} \\
	\hline
    \hline
    1 & Ephemeral & 22.10\% & - & - & - & 0.01\% & 0.01\% & 0.13\% & 0.44\% & 0.01\% & - \\
     & Intermittent & 62.75\% & 27.23\% & 49.67\% & 72.28\% & 28.87\% & 80.99\% & 32.86\% & 86.50\% & 72.05\% & 39.43\% \\
     & Perennial & 15.15\% & 72.77\% & 50.33\% & 27.72\% & 71.12\% & 19.00\% & 67.01\% & 13.06\% & 27.94\% & 60.57\% \\
    \hline
    2 & Ephemeral & 21.62\% & 0.01\% & - & - & 0.02\% & 0.01\% & 0.04\% & 0.17\% & - & - \\
     & Intermittent & 58.90\% & 15.61\% & 35.49\% & 58.66\% & 21.65\% & 68.16\% & 27.42\% & 84.56\% & 69.38\% & 29.93\% \\
     & Perennial & 19.49\% & 84.39\% & 64.51\% & 41.34\% & 78.34\% & 31.83\% & 72.54\% & 15.28\% & 30.62\% & 70.07\% \\
    \hline
    3 & Ephemeral & 30.77\% & - & - & - & 0.01\% & 0.01\% & 0.02\% & 0.04\% & - & - \\
     & Intermittent & 47.36\% & 10.17\% & 21.87\% & 38.50\% & 14.43\% & 45.42\% & 19.34\% & 74.35\% & 57.98\% & 16.41\% \\
     & Perennial & 21.87\% & 89.83\% & 78.13\% & 61.49\% & 85.56\% & 54.58\% & 80.64\% & 25.61\% & 42.02\% & 83.59\% \\
    \hline
    4 & Ephemeral & 55.70\% & - & - & - & 0.03\% & - & 0.04\% & 0.03\% & - & - \\
     & Intermittent & 23.89\% & 3.30\% & 4.98\% & 14.62\% & 9.85\% & 40.47\% & 13.19\% & 36.62\% & 26.80\% & 10.91\% \\
     & Perennial & 20.41\% & 96.70\% & 95.02\% & 85.38\% & 90.12\% & 59.53\% & 86.77\% & 63.35\% & 73.20\% & 89.09\% \\
    \hline
    5 & Ephemeral & 27.39\% & - & - & - & - & - & - & - & - & - \\
     & Intermittent & 8.05\% & - & - & 2.93\% & - & - & - & - & 6.30\% & - \\
     & Perennial & 64.56\% & - & 100.0\% & 97.07\% & 100.0\% & - & 100.0\% & 100.0\% & 93.70\% & - \\
    \hline
    All & Ephemeral & 22.80\% & - & - & - & 0.01\% & 0.01\% & 0.09\% & 0.31\% & 0.01\% & - \\
     Stream & Intermittent & 59.86\% & 22.34\% & 42.45\% & 64.51\% & 24.88\% & 73.73\% & 29.76\% & 84.92\% & 69.67\% & 33.96\% \\
     Orders & Perennial & 17.34\% & 77.66\% & 57.55\% & 35.49\% & 75.10\% & 26.27\% & 70.15\% & 14.77\% & 30.32\% & 66.04\% \\
     \hline
\end{NiceTabular}
 \def\tablename{Supplementary Table}
    
    \vspace{1em}   
    \centering
\begin{NiceTabular}{|c|c||c|c|c|c|c|c|c|c|c|}
	\hline
	\textbf{\makecell{Stream\\Order}} & \textbf{\makecell{Hydrographic\\Category}} & \textbf{\makecell{HU2\\10}} & \textbf{\makecell{HU2\\11}} & \textbf{\makecell{HU2\\12}} & \textbf{\makecell{HU2\\13}} & \textbf{\makecell{HU2\\14}} & \textbf{\makecell{HU2\\15}} & \textbf{\makecell{HU2\\16}} & \textbf{\makecell{HU2\\17}} & \textbf{\makecell{HU2\\18}} \\
    \hline
    \hline
    1 & Ephemeral & 3.06\% & 1.73\% & - & 68.67\% & 6.81\% & 82.20\% & 94.44\% & 2.68\% & 89.30\% \\
     & Intermittent & 89.61\% & 88.54\% & 95.07\% & 30.57\% & 88.67\% & 17.31\% & 2.68\% & 77.57\% & 8.69\% \\
     & Perennial & 7.32\% & 9.73\% & 4.93\% & 0.76\% & 4.52\% & 0.49\% & 2.88\% & 19.74\% & 2.02\% \\
    \hline
    2 & Ephemeral & 1.29\% & 0.52\% & - & 70.04\% & 1.88\% & 82.26\% & 92.67\% & 1.75\% & 84.40\% \\
     & Intermittent & 90.01\% & 88.78\% & 94.80\% & 28.79\% & 90.72\% & 17.23\% & 3.25\% & 66.42\% & 12.16\% \\
     & Perennial & 8.70\% & 10.70\% & 5.20\% & 1.17\% & 7.40\% & 0.51\% & 4.08\% & 31.83\% & 3.45\% \\
    \hline
    3 & Ephemeral & 0.84\% & 0.25\% & - & 67.17\% & 0.77\% & 78.82\% & 93.38\% & 1.17\% & 82.73\% \\
     & Intermittent & 88.35\% & 85.50\% & 92.41\% & 31.76\% & 90.34\% & 20.66\% & 3.01\% & 54.82\% & 14.03\% \\
     & Perennial & 10.80\% & 14.25\% & 7.59\% & 1.07\% & 8.89\% & 0.51\% & 3.61\% & 44.01\% & 3.24\% \\
    \hline
    4 & Ephemeral & 0.74\% & 0.04\% & - & 61.15\% & 0.60\% & 76.48\% & 97.80\% & 0.76\% & 89.00\% \\
     & Intermittent & 84.17\% & 76.38\% & 71.05\% & 38.29\% & 89.50\% & 22.95\% & 1.13\% & 54.91\% & 8.95\% \\
     & Perennial & 15.09\% & 23.58\% & 28.95\% & 0.55\% & 9.90\% & 0.57\% & 1.07\% & 44.33\% & 2.06\% \\
    \hline
    5 & Ephemeral & - & - & - & 16.07\% & - & 56.35\% & 97.62\% & - & 97.03\% \\
     & Intermittent & - & - & 16.98\% & 83.93\% & 100.0\% & 29.63\% & 0.75\% & 10.67\% & 2.30\% \\
     & Perennial & 100.0\% & - & 83.02\% & - & - & 14.02\% & 1.63\% & 89.33\% & 0.67\% \\
    \hline
    All & Ephemeral & 2.26\% & 1.17\% & - & 68.96\% & 4.43\% & 81.72\% & 93.66\% & 2.22\% & 86.89\% \\
     Stream & Intermittent & 89.67\% & 88.38\% & 94.76\% & 30.10\% & 89.58\% & 17.78\% & 2.93\% & 71.63\% & 10.48\% \\
     Orders & Perennial & 8.08\% & 10.45\% & 5.24\% & 0.94\% & 5.99\% & 0.50\% & 3.41\% & 26.15\% & 2.63\% \\
     \hline
    \end{NiceTabular}
     \def\tablename{Supplementary Table}
    \caption{The percentage of our waterway points with a label of Ephemeral, Intermittent, or Perennial for each stream order and HU2 region. See figure \ref{fig:hu2} for a map of the HU2 regions.} 
    \label{tab:streamtype}
\end{table}

\begin{table}
    \centering

\begin{NiceTabular}{|c||c|c||c|c||c|c|}
    \hline
    \textbf{\makecell{fcode Description}} & \textbf{\makecell{Count\\(Model)}} & \textbf{\makecell{Percent\\(Model)}} & \textbf{\makecell{Count\\(TDX)}} & \textbf{\makecell{Percent\\(TDX)}} & \textbf{\makecell{Count\\(All)}} & \textbf{\makecell{Percent\\(All)}} \\
    \hline
    \hline
    Area of Complex Channels &  62,740 & 0.01\% &  166,091 & 0.05\% &  228,831 & 0.03\% \\
    \hline
    Bay/Inlet &  69,502 & 0.02\% &  45,063 & 0.01\% &  114,565 & 0.02\% \\
    \hline
    Canal/Ditch &  4,692,647 & 1.03\% &  7,451,066 & 2.45\% &  12,143,713 & 1.60\% \\
    \hline
    Coastline &  1,225,997 & 0.27\% &  339,909 & 0.11\% &  1,565,906 & 0.21\% \\
    \hline
    Connector &  785,219 & 0.17\% &  887,295 & 0.29\% &  1,672,514 & 0.22\% \\
    \hline
    Dam/Weir &  125,658 & 0.03\% &  93,355 & 0.03\% &  219,013 & 0.03\% \\
    \hline
    Drainageway &  491,065 & 0.11\% &  801,105 & 0.26\% &  1,292,170 & 0.17\% \\
    \hline
    Estuary &  1,373,176 & 0.30\% &  462,242 & 0.15\% &  1,835,418 & 0.24\% \\
    \hline
    Foreshore &  136,968 & 0.03\% &  51,813 & 0.02\% &  188,781 & 0.02\% \\
    \hline
    Inundation Area &  602,995 & 0.13\% &  866,319 & 0.29\% &  1,469,314 & 0.19\% \\
    \hline
    Lake/Pond &  5,103,630 & 1.12\% &  3,004,694 & 0.99\% &  8,108,324 & 1.07\% \\
    \hline
    Lake/Pond: Intermittent &  1,399,406 & 0.31\% &  1,770,533 & 0.58\% &  3,169,939 & 0.42\% \\
    \hline
    Lake/Pond: Perennial &  49,654,576 & 10.88\% &  24,681,554 & 8.12\% &  74,336,130 & 9.78\% \\
    \hline
    Levee &  45,756 & 0.01\% &  25,575 & 0.01\% &  71,331 & 0.01\% \\
    \hline
    Nonearthen Shore &  133,328 & 0.03\% &  102,021 & 0.03\% &  235,349 & 0.03\% \\
    \hline
    Pipeline &  230,606 & 0.05\% &  207,672 & 0.07\% &  438,278 & 0.06\% \\
    \hline
    Playa &  286,517 & 0.06\% &  713,250 & 0.23\% &  999,767 & 0.13\% \\
    \hline
    Rapids &  10,036 & 0.00\% &  112,762 & 0.04\% &  122,798 & 0.02\% \\
    \hline
    Reservoir &  1,485,625 & 0.33\% &  722,081 & 0.24\% &  2,207,706 & 0.29\% \\
    \hline
    Sea/Ocean &  702,021 & 0.15\% &  391,905 & 0.13\% &  1,093,926 & 0.14\% \\
    \hline
    Sounding Datum Line &  81,985 & 0.02\% &  49,286 & 0.02\% &  131,271 & 0.02\% \\
    \hline
    Stream/River &  7,117,477 & 1.56\% &  5,589,661 & 1.84\% &  12,707,138 & 1.67\% \\
    \hline
    Stream/River: Ephemeral &  62,111,529 & 13.61\% &  22,918,445 & 7.54\% &  85,029,974 & 11.19\% \\
    \hline
    Stream/River: Intermittent &  163,080,866 & 35.74\% &  80,295,191 & 26.42\% &  243,376,057 & 32.02\% \\
    \hline
    Stream/River: Perennial &  47,223,243 & 10.35\% &  100,498,353 & 33.07\% &  147,721,596 & 19.43\% \\
    \hline
    Swamp/Marsh &  7,435,657 & 1.63\% &  5,903,597 & 1.94\% &  13,339,254 & 1.75\% \\
    \hline
    Swamp/Marsh: Intermittent &  38,003 & 0.01\% &  38,792 & 0.01\% &  76,795 & 0.01\% \\
    \hline
    Swamp/Marsh: Perennial &  28,157 & 0.01\% &  42,139 & 0.01\% &  70,296 & 0.01\% \\
    \hline
    Underground Conduit &  23,781 & 0.01\% &  30,299 & 0.01\% &  54,080 & 0.01\% \\
    \hline
    Unknown &  100,176,619 & 21.96\% &  44,416,129 & 14.61\% &  144,592,748 & 19.02\% \\
    \hline
    Wash &  330,076 & 0.07\% &  1,234,797 & 0.41\% &  1,564,873 & 0.21\% \\
    \hline
\end{NiceTabular}
 \def\tablename{Supplementary Table}
    \caption{The percentage and count of each data sources points labels by waterway type}
    \label{tab:allfcodes}
\end{table}

\begin{table}[]
    \centering
    \begin{tabular}{|c|c|}
        \hline
        Water(way) Type & Weight \\
        \hline
        playa & 0.0 \\
        \hline
        Inundation area & 0.0 \\
        \hline
        Swamp Intermittent & 0.5 \\
        \hline
        Swamp Perennial & 0.5 \\
        \hline
        Swamp & 0.5 \\
        \hline
        Reservoir & 2. \\
        \hline
        Lake Intermittent & 0.5 \\
        \hline
        Lake Perennial & 7. \\
        \hline
        Lake & 7. \\
        \hline
        spillway & 0.0 \\
        \hline
        drainage & 0.5 \\
        \hline
        wash & 0.5 \\
        \hline
        canal storm & 0.5 \\
        \hline
        canal aqua & 1. \\
        \hline
        canal & 0.5 \\
        \hline
        artificial path & 1.0 \\
        \hline
        Ephemeral Streams & 7.5 \\
        \hline
        Intermittent Streams & 7.5 \\
        \hline
        Perennial Streams & 6.5 \\
        \hline
        Streams Other & 6.5 \\
        \hline
        other & 1. \\
        \hline
    \end{tabular}
    \def\tablename{Supplementary Table}
    \caption{Model weights for fcode labels. A weight of 0 indicates the NHD data were considered to not be waterways, a weight between 0 and 1 were masked out, weights greater than or equal to 1 were used to scale the BCE loss contribution of that pixel by that amount.
}
    \label{tab:edt2}
\end{table}

\begin{figure}
    \def\figurename{Supplementary Figure}
    \setcounter{figure}{0}
    \centering
    \includegraphics[width=6in]{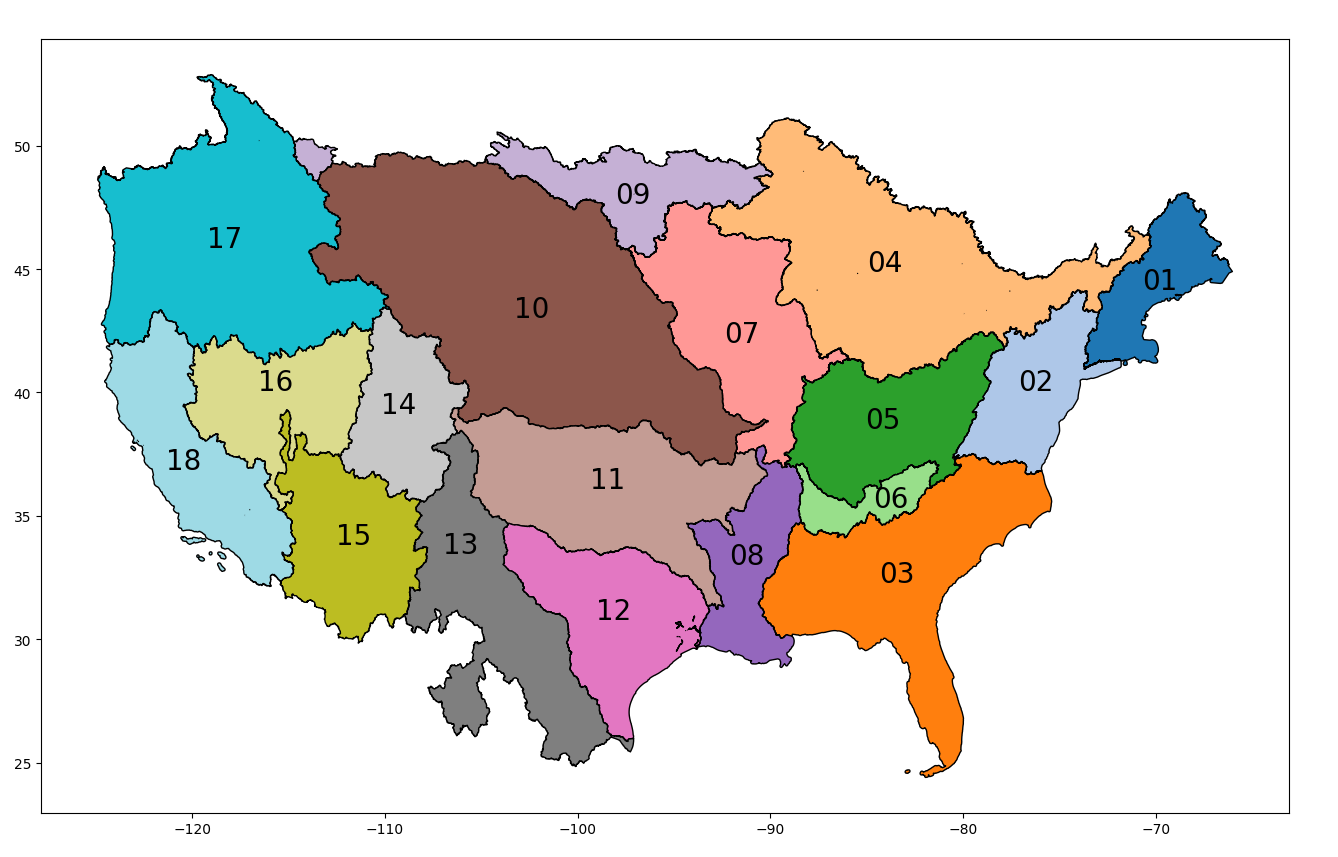}
  
    \caption{The HU2 Regions}
    \label{fig:hu2}
\end{figure}
\newpage

\begin{figure}[H]
  \def\figurename{Supplementary Figure}
    \centering
    \includegraphics[width=6in]{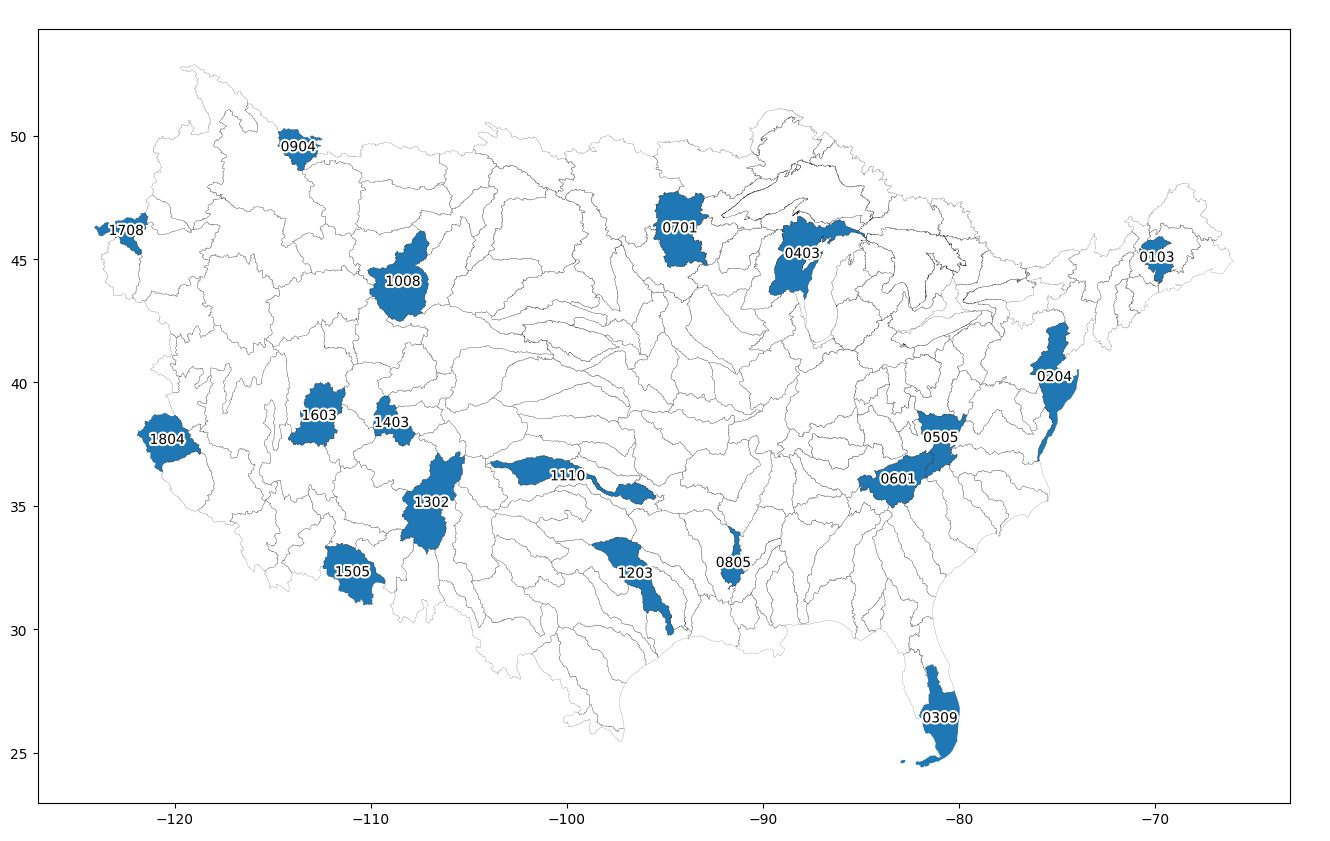}
    \small
    \caption{A map of the HU4 regions used for testing.}
    \label{fig:hu4}
\end{figure}

\newpage

\section{Model Tables} \label{modtab}
If input is a number, then that number refers to the layer in the same table with the corresponding layer number. All normalization layers are instance normalization.

\setcounter{table}{0}

\begin{table}[H]
    \centering
    \begin{NiceTabular}{|c|c|c|c|}
        \hline
        Layer Number & Layer & Inputs & Output sizes (channels, rows, columns)\\
        \hline
        \hline
        1 & Attention & - & $(2^4,\, R,\, C)$ \\
        \hline
        2 &  Encoder \ref{tab:encoder} & 1 & $(2^5,\, R/2^1,\, C/2^1)$\\
        \hline
        3 &  Encoder & 2 & $(2^6,\, R/2^2,\, C/2^2)$\\
        \hline
        4 &  Encoder & 3 & $(2^7,\, R/2^3,\, C/2^3)$\\
        \hline
        5 &  Encoder & 4 & $(2^8,\, R/2^4,\, C/2^4)$\\
        \hline
        6 &  Encoder & 5 & $(2^9,\, R/2^5,\, C/2^5)$\\
        \hline
        7 &  Decoder \ref{tab:decoder} & 6, 5 & $(2^8,\, R/2^4,\, C/2^4)$\\
        \hline
        8 &  Decoder & 7, 4 & $(2^7,\, R/2^3,\, C/2^3)$\\
        \hline
        9 &  Decoder & 8, 3 & $(2^6,\, R/2^2,\, C/2^2)$\\
        \hline
        10 &  Decoder & 9, 2 & $(2^5,\, R/2^1,\, C/2^1)$\\
        \hline
        11 &  Fully Connected & 10 & $(1,\, R/2^1,\, C/2^1)$\\
        \hline
    \end{NiceTabular}
    \def\tablename{Model Table}
        \caption{\textbf{Main Model Layers}: Original input size of $(Ch, R, C)$}
    \label{tab:model}
\end{table}

\begin{table}[H]
    \centering
    \begin{NiceTabular}{|c|c|c|c|}
    \hline
        Layer Number & Layer & Input & Output size\\
        \hline
        \hline
        1 & 2x2 Convolution with stride 2 & Previous Encoder & $(Ch,\, R/2,\, C/2)$ \\
        \hline
        2 & Normalization & 1 & $(Ch,\, R/2,\, C/2)$ \\
        \hline
        3 & Multiplication Block \ref{tab:multiplication} & 2 & $(Ch,\, R/2,\, C/2)$ \\
        \hline
        4 & Residual Block \ref{tab:residualB} & 2 & $(Ch,\, R/2,\, C/2)$ \\
        \hline
        5 & Normalization & 3 & $(Ch,\, R/2,\, C/2)$ \\
        \hline
        6 & Normalization & 4 & $(Ch,\, R/2,\, C/2)$ \\
        \hline
        7 & Concatenate & 2,5,6 & $(3Ch,\, R/2,\, C/2)$ \\
        \hline
        8 & Fully Connected & 7 & $(2Ch,\, R/2,\, C/2)$ \\
        \hline
    \end{NiceTabular}
    \def\tablename{Model Table}
    \caption{\textbf{Internal Encoder Layers}: Previous encoder output size of $(Ch,\, R,\, C)$}
    \label{tab:encoder}
\end{table}

\begin{table}[H]
    \centering
    \begin{NiceTabular}{|c|c|c|c|}
        \hline
        Layer Number & Layer & Input & Output size\\
        \hline
        \hline
        1 & 2x2 Transposed Convolution & Previous Decoder & $(Ch,\, R,\, C)$\\
        \hline
        2 & Normalization & 1 & $(Ch,\, R,\, C)$\\
        \hline
        3 & Concatenate & 2, Skip Connection & $(2Ch,\, R,\, C)$\\
        \hline
        4 & Multiplication Block \ref{tab:multiplication} & 3 & $(2Ch,\, R,\, C)$\\
        \hline
        5 & Residual Block \ref{tab:residualB} & 3 & $(2Ch,\, R,\, C)$\\
        \hline
        6 & Normalization & 4 & $(2Ch,\, R,\, C)$\\
        \hline
        7 & Normalization & 5 & $(2Ch,\, R,\, C)$\\
        \hline
        8 & Concatenate & 6, 7 & $(4Ch,\, R,\, C)$\\
        \hline
        9 & Fully Connected & 8 & $(2Ch,\, R,\, C)$\\
        \hline
        10 & Normalization & 9 & $(2Ch,\, R,\, C)$\\
        \hline
        11 & Convolution Block \ref{tab:convolution} & 10 & $(Ch,\, R,\, C)$\\
        \hline
        12 & Normalization & 11 & $(Ch,\, R,\, C)$\\
        \hline     
    \end{NiceTabular}
    \def\tablename{Model Table}
    \caption{\textbf{Internal Decoder Layers}: Previous decoder output size of $(2Ch,\, R/2,\, C/2)$ skip connection size of $(Ch,\, R,\, C)$}
    \label{tab:decoder}
\end{table}

\begin{table}[H]
    \centering
    \begin{NiceTabular}{|c|c|c|}
    \hline
        Layer Number & Layer & Input \\
        \hline
        \hline
        1 & Convolution & initial input\\
        \hline
        2 & Leaky ReLU & 1 \\
        \hline
        3 & Convolution & 2 \\
        \hline
        4 & Add & 3, initial input \\
        \hline
        5 & Normalization & 4 \\
        \hline
    \end{NiceTabular}
    \def\tablename{Model Table}
    \caption{\textbf{Residual Layer}: The convolutions were either $5\times5$ or $3\times3$, with zero padding.}
    \label{tab:residual}
\end{table}

\begin{table}[H]
    \centering
    \begin{NiceTabular}{|c|c|c|}
    \hline
        Layer Number & Layer & Input \\
        \hline
        \hline
        1 & Residual Layer \ref{tab:residual} & initial input\\
        \hline
        2 & Residual Layer & 1 \\
        \hline
        3 & Residual Layer & 2 \\
        \hline
    \end{NiceTabular}
    \def\tablename{Model Table}
    \caption{\textbf{Residual Block}}
    \label{tab:residualB}
\end{table}

\begin{table}[H]
    \centering
    \begin{NiceTabular}{|c|c|c|}
    \hline
        Layer Number & Layer & Input \\
        \hline
        \hline
        1 & Convolution & initial input\\
        \hline
        2 & Leaky ReLU & 1 \\
        \hline
        3 & Convolution & 2 \\
        \hline
    \end{NiceTabular}
    \def\tablename{Model Table}
    \caption{\textbf{Convolution Block}: The two convolutions were the same $n\times n$, with zero padding, where $n=1, 3, 5,$ or $7$}
    \label{tab:convolution}
\end{table}

\begin{table}[H]
    \centering
    \begin{NiceTabular}{|c|c|c|}
    \hline
        Layer Number & Layer & Input \\
        \hline
        \hline
        1 & Convolution Block \ref{tab:convolution} & initial input\\
        \hline
        2 & Convolution Block & 1 \\
        \hline
        3 & Convolution Block & 2 \\
        \hline
        4 & Multiply & 3, initial input \\
        \hline
    \end{NiceTabular}
    \def\tablename{Model Table}
    \caption{\textbf{Multiplication Block}: Similar to a GLU, except we don't apply a second transformation to the initial input before multiplying it by the transformed input. Here, depending on how deep in the "U" the layer was, the first convolution block was $7\times7$, $5\times5$, or $3\times3$, the second was $5\times5$ or $3\times3$, and the final was $3\times3$}

    \label{tab:multiplication}
\end{table}

\newpage

\end{document}